\documentclass{article}
\usepackage{apacite}
\usepackage[preprint]{neurips_2023}

\usepackage[utf8]{inputenc} 
\usepackage[T1]{fontenc}    
\usepackage{hyperref}       
\usepackage{url}            
\usepackage{booktabs}       
\usepackage{amsfonts}       
\usepackage{nicefrac}       
\usepackage{microtype}      
\usepackage{xcolor}         

\title{Towards Machines that Trust: AI Agents Learn to Trust in the Trust Game}

\author{%
  Ardavan S.~Nobandegani\\
  Mila -- Quebec AI Institute \\
  Department of Psychology, McGill University \\
  Montreal, QC, Canada \\
  \texttt{ardavan.salehinobandegani@mcgill.ca} \\
  \And
  Irina Rish \\
  Mila -- Quebec AI Institute \\
  Montreal, QC, Canada \\
  \texttt{irina.rish@mila.quebec} \\
  \And
  Thomas R.~Shultz \\
  School of Computer Science, McGill University \\
  Department of Psychology, McGill University \\
  Montreal, QC, Canada \\
  \texttt{thomas.shultz@mcgill.ca} \\
}

\usepackage{algorithm}
\usepackage{algorithmic}

\usepackage{amsmath}
\DeclareMathOperator*{\argmax}{argmax}

\usepackage{amsfonts}
\usepackage{MnSymbol}

\usepackage{color}

\usepackage{amsthm}
\usepackage{amsmath}
\usepackage{booktabs}
\usepackage{mathtools}
\usepackage{multirow}
\usepackage{graphicx}
\usepackage{cleveref}
\usepackage{svg}
\usepackage{subcaption}

\usepackage{xspace}
\newcommand{\Trustor}{\texttt{Trustor}\xspace}
\newcommand{\Trustee}{\texttt{Trustee}\xspace}

\begin{document}
\maketitle

\begin{abstract}
Widely considered a cornerstone of human morality, trust shapes many aspects of human social interactions. In this work, we present a theoretical analysis of the \textit{trust game}, the canonical task for studying trust in behavioral and brain sciences, along with simulation results supporting our analysis. Specifically, leveraging reinforcement learning (RL) to train our AI agents, we systematically investigate learning trust under various parameterizations of this task. Our theoretical analysis, corroborated by the simulations results presented, provides a mathematical basis for the emergence of trust in the trust game.
\end{abstract}

\section{Introduction}
Widely viewed as a bedrock of human morality \citep{mitkidis2017trust}, trust proves to be essential for virtually all forms of human interactions, from love and friendship to economic growth and the emergence of large-scale organizations \citep[e.g.,][]{uysal2012reciprocal,slovic1993perceived,mayer1995integrative,gachter1998effective,hetherington1998political,fehr2003detrimental,campbell2019adult}. 

In this work, we present a theoretical analysis of the \textit{trust game} \citep{berg1995trust}, the canonical task for studying trust in behavioral and brain sciences \citep[e.g.,][]{camerer1988experimental,fehr2009economics,tzieropoulos2013trust,johnson2011trust,mitkidis2017trust,alos2019trust}. Additionally, leveraging reinforcement learning (RL) to train our AI agents, we systematically investigate learning trust under various parameterizations of this task. As we show, our simulation results support our theoretical analysis.

\paragraph{The trust game (TG):} Two players, \texttt{Trustor} and \texttt{Trustee}, are anonymously paired. \texttt{Trustor} is given a monetary endowment $T$ and should choose what fraction $r\in[0,1]$ of it will be sent to \texttt{Trustee}, as an indication of trust. The transferred amount $rT$ is then multiplied by a factor $K>0$ and received by \texttt{Trustee}. Finally, \texttt{Trustee} should decide what fraction $\alpha\in[0,1]$ of their endowment $KrT$ they are willing to return to \texttt{Trustor}. 

\section{Theoretical Analysis}
Here we present a theoretical analysis of TG, providing a mathematical basis for the emergence of trust in that game.

\paragraph{Theorem 1.} \textit{Assuming that \emph{\texttt{Trustee}} returns $\alpha(r)\in[0,1]$ fraction of their endowment to \emph{\texttt{Trustor}} with probability $p(r)\in[0,1]$ and returns nothing otherwise, the optimal fraction $r^\ast\in[0,1]$ (in the sense of maximizing \emph{\texttt{Trustor}'s} expected reward) that \emph{\texttt{Trustor}} should transfer to \emph{\texttt{Trustee}} is given by:}
\begin{equation}
\label{theorem_1}
r^\ast = \argmax_{r\in[0,1]}\,\, (\alpha(r)p(r)K-1)r.
\end{equation}
 \emph{Note that the optimal fraction may not be unique, in which case $\argmax$ in \emph{(\ref{theorem_1})} returns a set.}

\paragraph{Proposition 1.} \textit{Let $\alpha_0,p_0\in[0,1]$ and $m,n\in\mathbb{N} \cup \{0\}$. Assuming that \emph{\texttt{Trustee}} returns $\alpha(r)=\alpha_0r^m$ fraction of their endowment to \emph{\texttt{Trustor}} with probability $p(r)=p_0r^n$ and returns nothing otherwise, the optimal fraction $r^\ast$ (in the sense of maximizing \emph{\texttt{Trustor}'s} expected reward) that \emph{\texttt{Trustor}} should transfer to \emph{\texttt{Trustee}} is given by:}
\begin{equation}
r^\ast = 
    \begin{cases}
      1  & \text{if $\alpha_0 p_0 K > 1$}\\
      0  & \text{if $\alpha_0 p_0 K < 1$}.
    \end{cases}
 \end{equation}

According to Proposition 1, if $\alpha_0 p_0 K > 1$, \Trustor should show {complete trust} by transferring their whole monetary endowment $T$ to \Trustee. Conversely, if $\alpha_0 p_0 K < 1$, \Trustor should exhibit no trust by transferring nothing to \Trustee.

\section{RL \Trustor Learns to Trust}
Here we leverage RL to train \Trustor under the broad assumptions of Theorem 1. Concretely, we have RL \Trustor learn about trust by interacting with a TG \Trustee that behaves in accord with the assumptions of Theorem 1.

\begin{figure*}[!ht]
\centering
\includegraphics[trim = 0mm 20pt 0mm 0mm,clip,width=1\textwidth]{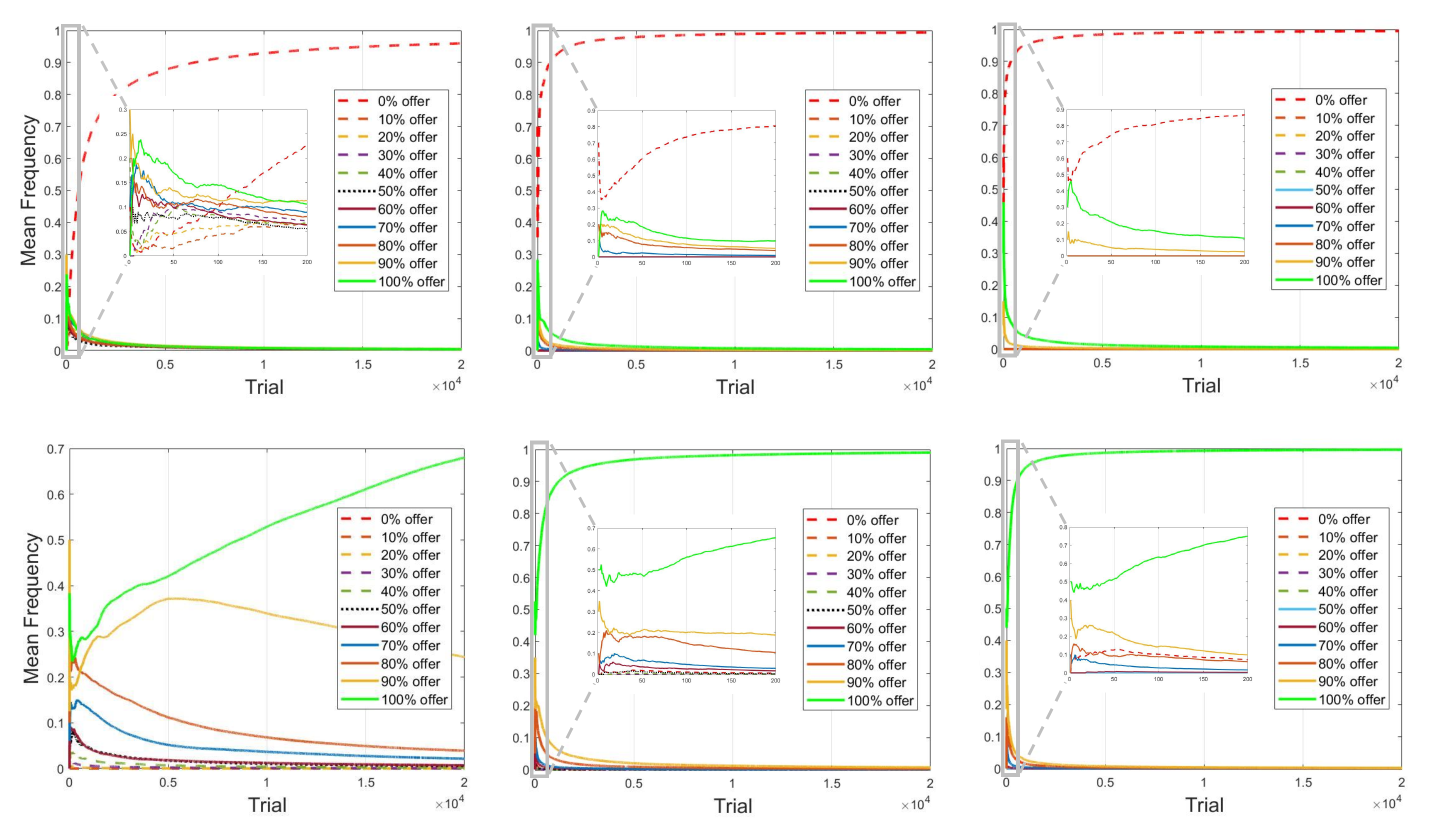}
\caption{Mean frequency of RL \Trustor's transfers, with $\alpha(r)=\alpha_0 r^m$ and $p(r)=p_0 r^n$; see Proposition 1. {(top row)} Simulating the $\alpha_0p_0K<1$ condition ($\alpha_0=0.5, p_0=0.5, K=3$) when $\alpha(r)$ and $p(r)$ are:  (left) constant ($m=n=0$), (middle) linear ($m=n=1$), or (right) quadratic ($m=n=2$). {(bottom row)} Simulating the $\alpha_0p_0K>1$ condition ($\alpha_0=1, p_0=0.5, K=3$) when $\alpha(r)$ and $p(r)$ are:  (left) constant ($m=n=0$), (middle) linear ($m=n=1$), or (right) quadratic ($m=n=2$). As a visual aid, the dynamics for the first 200 trials are provided in a smaller plot, located at the center of each subplot. Note that these simulation results are fully consistent with Proposition 1, supporting our mathematical analysis.}
\label{fig_trust}
\end{figure*}

To train RL \Trustor, we adopt Thompson Sampling  \citep{thompson1933likelihood}, a well-established method that enjoys near-optimality guarantees \citep{agrawal2012analysis,agrawal2013further}. The RL \Trustor should decide what fraction of their total endowment $T$ they are willing to transfer to TG \Trustee. For ease of analysis, here we assume that RL \Trustor chooses between a finite set of actions (fractions): $\mathcal{A} = \{0, 0.1, 0.2, \ldots, 0.9, 1\}$; see Algorithm 1.

\begin{algorithm}
\caption{{Thompson Sampling for RL \Trustor}}\label{TG_trustor}
\begin{algorithmic}[2]
\STATE \textbf{Initialize}. $\forall r\in\mathcal{A}$: $S_{r} = 0$ and $F_{r} = 0$\newline
1:\hspace*{3pt} \textbf{for} $i = 1,\ldots, N$\newline
2:\hspace*{10pt}\hspace*{3pt} $\forall r\in\mathcal{A}$ compute:

\hspace*{10pt}\hspace*{10pt} $s_r = (T-rT+KrT\alpha(r))\beta_r + (T-rT)(1-\beta_r)$, where $\beta_r\sim$ Beta$(S_{r}+1, F_{r}+1)$

3:\hspace*{10pt}\hspace*{3pt} $r^\filledstar = \argmax_{r} s_{r}$ and  $u\sim U[0,1]$\newline
4:\hspace*{10pt}\hspace*{3pt} \textbf{if} $u<p(r^\filledstar)$ \textbf{then}\newline
5:\hspace*{10pt}\hspace*{0pt} \hspace*{20pt}$S_{r^\filledstar} = S_{r^\filledstar} + 1$\newline
6:\hspace*{10pt}\hspace*{3pt} \textbf{else}\newline
7:\hspace*{10pt}\hspace*{0pt} \hspace*{20pt}$F_{r^\filledstar} = F_{r^\filledstar} + 1$\newline
8:\hspace*{10pt}\hspace*{3pt} \textbf{end if}\newline
\hspace*{-5pt}10:\hspace*{3pt} \textbf{end for}
\end{algorithmic}
\end{algorithm}

In Algorithm \ref{TG_trustor}, $S_r$ denotes the number of times \Trustee has returned a positive amount upon \Trustor
transferring a fraction $r\in\mathcal{A}$ of their endowment, and $F_r$ denotes the number of times \Trustee has returned a zero amount upon \Trustor
transferring a fraction $r$ of their endowment. Also, $U[0,1]$ denotes a uniform distribution with support $[0,1]$ and Beta$(\cdot,\cdot)$ is the Beta distribution.

Algorithm \ref{TG_trustor} can be described in simple terms as follows. Recall that \Trustor transfers $rT$, $r\in[0,1]$, to \Trustee who, with probability $1-p(r)$, returns a zero amount, and, with probability $p(r)$, returns $KrT\alpha(r)$ to \Trustor. At the start, i.e., prior to any learning, $S_r$ and $F_r$ are both set to zero. In each trial (for a total of $N$ trials), an estimate of expected reward for each transferred fraction $r\in\mathcal{A}$ is computed by sampling from the corresponding distribution (Line 2), and the fraction currently yielding the highest expected reward estimate $r^\filledstar$ is then chosen by RL \Trustor to be transferred to TG \Trustee (Line 3). Upon this transfer, TG \Trustee either returns a positive amount to \Trustor, in which case $S_{r^\filledstar}$ is incremented by one (Line 5), or returns a zero amount to \Trustor, in which case $F_{r^\filledstar}$ is incremented by one (Line 7). Note that the former happens with probability $p(r^\filledstar)$ and the latter with probability $1-p(r^\filledstar)$.

In Figure \ref{fig_trust}, we simulate 10 RL \Trustor{}s and report the mean frequency of a fraction being transferred to TG \Trustee over the past trials, for a total of $N=20,000$ trials. Note that the simulation results reported in Figure \ref{fig_trust} support our mathematical analysis in Proposition 1. As Proposition 1 indicates, if $\alpha_0p_0K<1$, the optimal strategy for \Trustor is to transfer nothing to \Trustee (i.e., $r^\ast=0$); Figure 1({top row}) is consistent with this result: as Figure 1({top row}) shows, the RL \Trustor eventually arrives at the decision that they should transfer nothing to \Trustee.  Also, according to Proposition 1, if $\alpha_0p_0K>1$, the optimal strategy for \Trustor is to transfer all of their endowment (i.e., $r^\ast=1$); Figure 1({bottom row}) is consistent with this result: as Figure 1({bottom row}) shows, the RL \Trustor eventually arrives at the decision that they should transfer all of their endowment to \Trustee.

\section{Conclusion}
Trust is a backbone of human morality. In this work, we presented a theoretical analysis of the \textit{trust game} \citep{berg1995trust}, the canonical task for studying trust in behavioral and brain sciences \citep{mitkidis2017trust,alos2019trust}. Also, leveraging reinforcement learning (RL), we systematically investigated
learning trust under various parameterizations of this task. Our simulation results fully support our mathematical analysis. Extending our recent work \citep{nobandegani2023ILHF}, the work presented here shows that using models of interacting agents (here, TG \Trustee) makes the training process safer, cheaper, and faster (here, RL \Trustor training).

\bibliographystyle{apacite}
\bibliography{trust_ref}
\end{document}